\documentclass[letterpaper]{article} % DO NOT CHANGE THIS
\usepackage{aaai2027}  % DO NOT CHANGE THIS
\usepackage[hyphens]{url}  % DO NOT CHANGE THIS
\usepackage{graphicx} % DO NOT CHANGE THIS
\usepackage{natbib}  % DO NOT CHANGE THIS AND DO NOT ADD ANY OPTIONS TO IT
\usepackage{caption} % DO NOT CHANGE THIS AND DO NOT ADD ANY OPTIONS TO IT
\usepackage{algorithm}
\usepackage{algorithmic}

\usepackage{algorithmic}
\usepackage{booktabs}       % professional-quality tables
\usepackage{amsfonts}       % blackboard math symbols
\usepackage{nicefrac}       % compact symbols for 1/2, 
\usepackage{amsthm,amsmath,amssymb}
\usepackage{microtype}      % microtypography
\usepackage{xcolor}         % colors
\usepackage{amsmath} 
\usepackage{times}
\usepackage{soul}
\usepackage{url}
\usepackage[utf8]{inputenc}
\usepackage{makecell}
\usepackage{pifont}
\usepackage{boldline}
\usepackage[table]{xcolor}
\newcommand{\cmark}{\ding{51}}
\newcommand{\xmark}{\ding{55}}
\usepackage{amsthm}
\usepackage{booktabs}
\usepackage{algorithm}
\usepackage{algorithmic}
\usepackage[switch]{lineno}
\usepackage{xcolor}
\usepackage{hhline}
\usepackage{boldline}
\usepackage{colortbl}
\usepackage{amsfonts}
\usepackage{pifont}
\usepackage{amsmath}
\usepackage{booktabs}
\usepackage{multirow}
\usepackage{array}
\definecolor{lightred}{RGB}{255,153,153} % 自定义浅红色

\usepackage{newfloat}
\usepackage{listings}
\DeclareCaptionStyle{ruled}{labelfont=normalfont,labelsep=colon,strut=off} % DO NOT CHANGE THIS
\floatstyle{ruled}
\newfloat{listing}{tb}{lst}{}
\floatname{listing}{Listing}

\usepackage{booktabs}
\definecolor{cvprblue}{rgb}{0.21,0.49,0.74}
\nocopyright 
\title{Is There Really a Camouflaged Object? Towards Realistic \\ Camouflaged Object Detection}
\author{
    Huafeng Chen\textsuperscript{\rm 1}
    Yueming Lyu\textsuperscript{\rm 1},
    Chenyang Si\textsuperscript{\rm 1},
    Wende Tan\textsuperscript{\rm 2},
    Liucheng Guo\textsuperscript{\rm 2},
    Caifeng Shan\textsuperscript{\rm 1}
}
\affiliations{
    \textsuperscript{\rm 1}School of Intelligence Science and Technology, Nanjing University\\
    \textsuperscript{\rm 2}Department of Computing, Imperial College London\\
}

\begin{document}

\maketitle

\begin{abstract}
Camouflaged object detection (COD) aims to segment objects that are visually concealed in their surroundings and has attracted increasing attention in recent years. However, most existing COD methods are developed under a closed-world assumption, where each input image is assumed to contain a camouflaged object. This assumption ignores realistic scenarios with pure backgrounds or non-camouflaged objects, causing existing models to produce severe false positives when deployed in open-world environments. To address this limitation, we propose OPC16K, a large-scale benchmark for realistic COD. OPC16K contains 16,245 images from 14 sources and is carefully organized into camouflaged-object images, pure background images, and non-camouflaged-object images, enabling comprehensive evaluation of both segmentation quality and negative-sample rejection. Based on this benchmark, we further propose OPCNet, a presence-aware camouflage network that reformulates COD from a pure segmentation task into a joint problem of object localization and camouflage existence reasoning. Specifically, OPCNet introduces hierarchical existence reasoning to distinguish CO, BG, and NOCOD scenarios, similarity-aware camouflage relation modeling to capture foreground-background camouflage cues, and existence-aware feature refinement to regulate segmentation features with existence predictions.  Extensive experiments on OPC16K demonstrate that OPCNet achieves superior performance under the proposed realistic COD evaluation protocol, significantly reducing false positives on negative samples while maintaining accurate camouflaged-object segmentation. \textit{Code and dataset will be released at https://github.com/2231122/OPCOD.}
\end{abstract}


\section{Introduction}

Camouflaged Object Detection (COD)~\cite{fan2020camouflaged,fan2021concealed,pang2022zoom} aims to identify objects concealed within their surrounding environments, with significant applications in search and rescue~\cite{fan2021concealed}, species discovery, medical image analysis~\cite{fan2020inf}, and military reconnaissance. In recent years, deep learning-based COD methods have achieved remarkable progress in segmentation accuracy.

However, a fundamental gap exists between current research paradigms and real-world deployment: existing models~\cite{he2025run,Liu_2025_ICCV} are trained and evaluated under the closed-world assumption that camouflaged objects are always present in the input image, completely neglecting negative-sample scenarios.  
This assumption introduces a severe bias, as in real-world deployment, models tend to produce excessive false positives when presented with pure background images (BG) or images containing non-camouflaged objects (NOCOD).
As shown in Tab.\ref{tab:1}, state-of-the-art COD methods RUN~\cite{he2025run} and USCNet~\cite{zhou2025rethinking} exhibit false positive rates of 97.4\% and 49.3\%, respectively, on our OPC16K test set. 
% In practice, false positives can incur significant costs in applications such as military reconnaissance, search and rescue, and medical imaging. More importantly, unlike SOD where images are intentionally captured with salient targets, realistic COD scenarios contain a large proportion of negative samples, making reliable negative-sample rejection essential for practical deployment.
In practice, false positives can severely compromise COD reliability in applications such as military reconnaissance, search and rescue, and medical imaging. Unlike SOD where images are intentionally captured with salient targets, realistic COD scenarios contain abundant negative samples, making reliable object-existence reasoning crucial for practical deployment.

\setlength{\tabcolsep}{10pt}
\begin{table}[t]
\caption{False positive rates of existing COD methods on the OPC16K test set. BG denotes the proportion of pure background images for which a method incorrectly predicts camouflaged objects, while NOCOD denotes the proportion of non-camouflaged object images for which a method produces false camouflaged-object predictions. AVG reports the average false positive rate over BG and NOCOD.}
    \label{tab:1}
    \vspace{-1mm}
\renewcommand{\arraystretch}{1}
    \centering
    % \vspace{-0.6em}
    \centering
    \scriptsize
     \begin{tabular}{lccc}
       \hlineB{2.5}
       \multirow{2}{*}{\textbf{Methods}}& \multicolumn{3}{c}{\textbf{OPC16K}}\\
    \cline{2-4}
        &\multicolumn{1}{c}{\textbf{BG$\downarrow$}} & \textbf{NOCOD$\downarrow$} & \multicolumn{1}{c}{\textbf{AVG$\downarrow$}}  \\
       \hlineB{2}
        
        CamoDiff~\cite{sun2025conditional}&0.953&1.00&0.977\\
        RUN~\cite{he2025run}&0.948&1.00&0.974\\USCNet~\cite{zhou2025rethinking}&0.202&0.784&0.493\\
       \hlineB{2.5}
    \end{tabular}
    
    \vspace{-2mm}
\end{table}

\setlength{\tabcolsep}{6pt}
\begin{table*}[!t]
\caption{Comparison of existing COD benchmarks. $^\dagger$: images included without mask annotations. NOCOD (Salient) and NOCOD (General) denote salient and general non-camouflaged object images, respectively. Balanced Sampling indicates whether negative samples consider the foreground-category and scene-level distributions of positive samples. Out-of-Domain Negatives denotes NOCOD samples whose categories are outside the camouflaged-object distribution.}
    \label{tab:benchmark_comparison}
    \vspace{-1mm}
\renewcommand{\arraystretch}{1.08}
    \centering
    \scriptsize
    \begin{tabular}{l|c|c|c|c|c|c|c|c|c}
       \hlineB{2.5}
       \textbf{Benchmark} 
       & \textbf{Source} 
       & \textbf{Target} 
       & \makecell{\textbf{\#Ann.}\\\textbf{IMG.}} 
       & \makecell{\textbf{COD}} 
       & \makecell{\textbf{BG}} 
       & \makecell{\textbf{NOCOD}\\\textbf{(Salient)}} 
       & \makecell{\textbf{NOCOD}\\\textbf{(General)}} 
       & \makecell{\textbf{Balanced}\\\textbf{Sampling}} 
       & \makecell{\textbf{Out-of-Domain}\\\textbf{Negatives}} \\
       \hlineB{2}
       CAMO~\cite{le2019anabranch} 
       & Internet & Conventional COD
       & 1,250 & 1,250 & \xmark & \xmark & \xmark & -- & -- \\
       
       CHAMELEON 
       & Internet & Conventional COD 
       & 76 & 76 & \xmark & \xmark & \xmark & -- & -- \\
       
       NC4K~\cite{lv2021simultaneously} 
       & Internet & Conventional COD 
       & 4,121 & 4,121 & \xmark & \xmark & \xmark & -- & -- \\
       
       COD10K~\cite{fan2020camouflaged} 
       & Internet & Conventional COD 
       & 7,000 & 5,066 & 1,934 & 3,000$^\dagger$ & \xmark & \xmark & \xmark \\
       
       USC12K~\cite{zhou2025rethinking} 
       & Filter \& Internet & Co-SOD\&COD 
       & 12,000 & 3,000 & 3,000 & 3,000 & \xmark & \xmark & 1,754/3,000 \\
       
       \hlineB{2}
       \rowcolor{cyan!10}
       \textbf{OPC16K (Ours)} 
       & \textbf{Filter \& Internet} & \textbf{Realistic COD} 
       & \textbf{16,245} & \textbf{9,000} & \textbf{3,050} & \textbf{1,076} & \textbf{3,119} & \cmark & \xmark \\
       \hlineB{2.5}
    \end{tabular}
    
    \vspace{-1mm}
\end{table*}

The most closely related work to ours is USCNet~\cite{zhou2025rethinking}, which targets the joint discrimination of salient objects (SO) and camouflaged objects (CO) in unconstrained scenes. However, its primary focus is on jointly modeling the two visually extreme concepts, particularly in scenes where both SO and CO co-exist, and it does not provide a systematic training and evaluation framework specifically designed for COD negative samples.
To address these limitations, we propose a comprehensive research framework for realistic COD, spanning three dimensions: dataset construction, method design, and evaluation protocol.

On the data side, we construct OPC16K, a large-scale realistic COD benchmark comprising over 16,000 images, including approximately 9,000 camouflaged-object samples and 7,000 negative samples (roughly 3,000 pure background images and 4,000 images containing non-camouflaged objects).
To reduce potential shortcut learning caused by distribution discrepancies, we carefully consider both foreground-category and scene-level distributions during negative-sample collection. This design aims to alleviate dataset biases and encourage models to focus more on camouflage-related visual cues rather than exploiting simple category or domain differences.
To our knowledge, OPC16K is the first dataset to systematically address the Realistic COD challenge.

On the method side, existing COD approaches~\cite{fan2021concealed,pang2022zoom,he2025run} formulate COD as a segmentation problem that assumes target existence by default. By relying mainly on pixel-wise segmentation supervision, these methods lack an explicit mechanism to distinguish whether the observed objects are truly camouflaged, leading to unreliable responses in BG and NOCOD scenarios. 
% To address this limitation, we propose OPCNet, which reformulates COD as an open-world discrimination framework involving three scenarios: COD, NOCOD, and BG.
To address this limitation, we propose OPCNet, which reformulates COD from a pure segmentation task into a joint framework of object localization and camouflage existence reasoning across three scenarios: COD, NOCOD, and BG.
OPCNet introduces explicit existence modeling and integrates three key mechanisms: 1) \textit{Hierarchical existence reasoning}, which decomposes the decision process into object presence verification and camouflage existence verification, enabling explicit discrimination among BG, NOCOD, and COD scenarios; 2) \textit{Existence-aware feature refinement}, which leverages hierarchical existence predictions to adaptively modulate segmentation features, bridging object localization and camouflage recognition; 3) \textit{Similarity-aware camouflage relation modeling}, which constructs foreground-background relational representations through prototype similarity and discrepancy modeling, providing explicit cues for camouflage existence reasoning under high foreground-background similarity.

On the evaluation side, existing metrics such as $S_m$ and MAE focus exclusively on segmentation quality for camouflaged positive samples and are incapable of measuring a model's holistic capability in realistic COD scenarios: the ability to suppress false positives on BG and NOCOD inputs, and the degree of segmentation degradation under negative-sample interference.
To this end, we propose a Realistic COD evaluation protocol that comprehensively assesses models along three dimensions: classification accuracy, negative-sample false positive rate, and segmentation-aware quality.
Our contributions are summarized as follows:

\begin{itemize}

\item We construct OPC16K, a large-scale benchmark specifically designed for realistic COD evaluation, with carefully controlled negative-sample distributions to assess reliable COD performance beyond the conventional positive-only assumption.

\item We propose OPCNet, which reformulates COD from a pure segmentation paradigm into a presence-aware discrimination framework through explicit existence modeling. By integrating hierarchical existence reasoning, existence-aware feature refinement, and similarity-aware camouflage relation modeling, OPCNet enables reliable COD under realistic COD scenarios.

\item We establish a comprehensive evaluation protocol for realistic COD evaluation and conduct extensive experiments on OPC16K, where OPCNet achieves superior performance across diverse evaluation metrics.
\end{itemize}

\section{Related Work}

\noindent\textbf{Camouflaged Object Detection.}
Early COD methods build upon encoder-decoder architectures to capture multi-scale features~\cite{fan2021concealed,fan2023advances,pang2022zoom,li2021uncertainty} and progressively refine segmentation maps~\cite{fan2020camouflaged,pang2022zoom,he2023weakly,guo2026boosting}. Subsequent works introduce auxiliary cues such as edge and texture awareness~\cite{he2023camouflaged,zhang2024unlocking,ye2025escnet} as well as multi-source inputs including depth maps and infrared imagery~\cite{wang2023depth,wang2024depth,Liu_2025_ICCV} to provide richer supervision signals. More recently, methods leveraging large foundation models such as SAM have demonstrated strong generalization in segmentation~\cite{chen2023sam,Liu_2025_ICCV,ren2025multi,chen2024just,SAM-COD,chen2026beyond}. Despite their advances in segmentation accuracy, all these methods are developed and evaluated under the closed-world assumption that every input image is guaranteed to contain a camouflaged object. This fundamentally limits their reliability in real-world deployment, where negative samples are prevalent and false positives carry significant practical costs.

\noindent\textbf{Salient Object Detection and Joint SOD/COD Methods.}
Salient Object Detection (SOD) aims to identify visually prominent objects that naturally attract human attention~\cite{zhao2019egnet}, standing in conceptual opposition to COD where objects are deliberately concealed. A line of works attempts to jointly train SOD and COD to achieve mutual enhancement, enabling a single model to handle both tasks~\cite{li2021uncertainty,luo2024vscode,zhao2024spider}. However, these methods still operate under the assumption that at least one foreground object is present in every image. USCNet~\cite{zhou2025rethinking} takes a step further by proposing a unified framework for joint discrimination of salient and camouflaged objects in unconstrained scenes, explicitly allowing both SO and CO to co-exist within a single image. Nevertheless, its primary focus remains on modeling the interplay between SO and CO, rather than systematically addressing the negative-sample challenge in COD. Our work is fundamentally different: we target the realistic COD setting and explicitly model the three-way discrimination among COD, NOCOD, and BG.

\noindent\textbf{COD Benchmarks and Datasets.}
Several datasets have been established to advance COD research, including CAMO~\cite{le2019anabranch}, COD10K~\cite{fan2020camouflaged}, and NC4K~\cite{lv2021simultaneously}. While these datasets have driven significant methodological progress, they are primarily constructed around camouflaged positive samples. Although CAMO and COD10K include a limited number of background and non-camouflaged images, these negative samples are scarce, category-wise homogeneous, and lack deliberate distribution alignment with positive samples in terms of foreground categories and background scenes. USC12K~\cite{zhou2025rethinking}, proposed alongside USCNet to support research on salient and camouflaged objects in unconstrained scenes, similarly includes SOD and background images as negatives. However, its non-camouflaged samples are limited to salient objects, and the background scene diversity remains constrained. Our proposed OPC16K addresses these limitations by incorporating a large and diverse set of negative samples whose scene and category distributions are deliberately matched to the positive samples, preventing models from exploiting domain discrepancies as shortcut solutions.

\begin{figure}[!t]
\centering
\includegraphics[width=\linewidth]{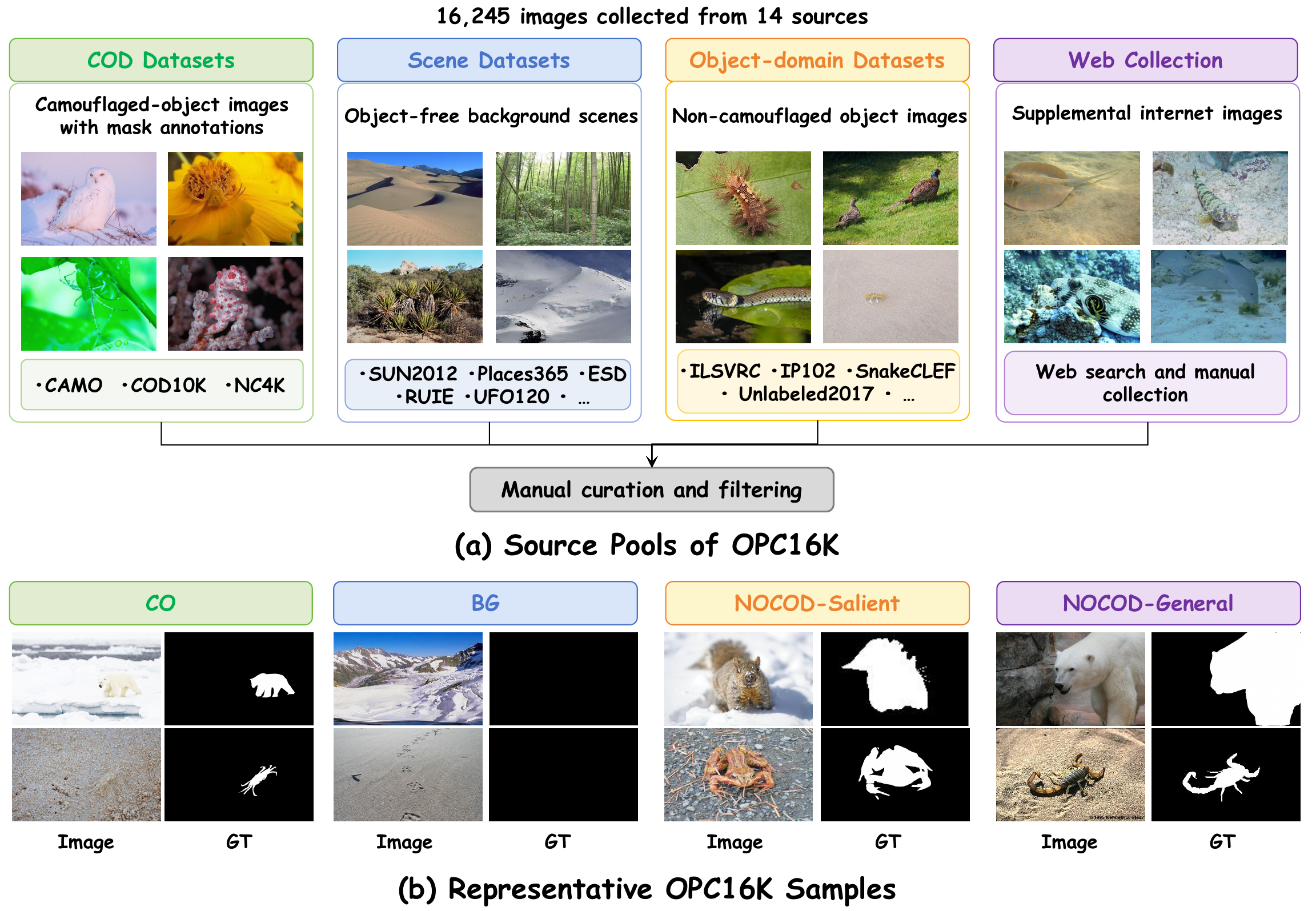}\\
\caption{Overview of OPC16K. OPC16K is constructed from four source pools, including COD datasets, scene datasets, object-domain datasets, and web collection. Representative image-mask pairs show the four visual scenarios in our benchmark: camouflaged-object images (CO), pure background images (BG), salient non-camouflaged-object images (NOCOD-Salient), and general non-camouflaged-object images (NOCOD-General).
}
\label{fig:1}
\vspace{-2mm}
\end{figure}

\section{The Proposed OPC16K Dataest}
Existing COD datasets, such as COD10K~\cite{fan2020camouflaged}, CAMO~\cite{le2019anabranch}, and NC4K~\cite{lv2021simultaneously}, are constructed exclusively around camouflaged positive samples, encoding the positive-only assumption at the data level. Although CAMO and COD10K contain a limited number of background and non-camouflaged images, these negative samples are scarce, category-wise homogeneous, and lack deliberate distribution alignment with positive samples. This data-level bias prevents models from learning to recognize the absence of camouflage, and renders existing benchmarks inadequate for evaluating real-world COD performance. Therefore, we introduce OPC16K, a large-scale dataset designed to support realistic COD research beyond the conventional positive-only assumption. It includes camouflaged positive samples, pure background images, and non-camouflaged object images, with scene and category distributions carefully aligned across all three subsets to ensure that the challenge stems from camouflage itself rather than domain discrepancies. We describe the details of OPC16K in terms of three key aspects, as follows.

\begin{figure}[!t]
\centering
\includegraphics[width=\linewidth]{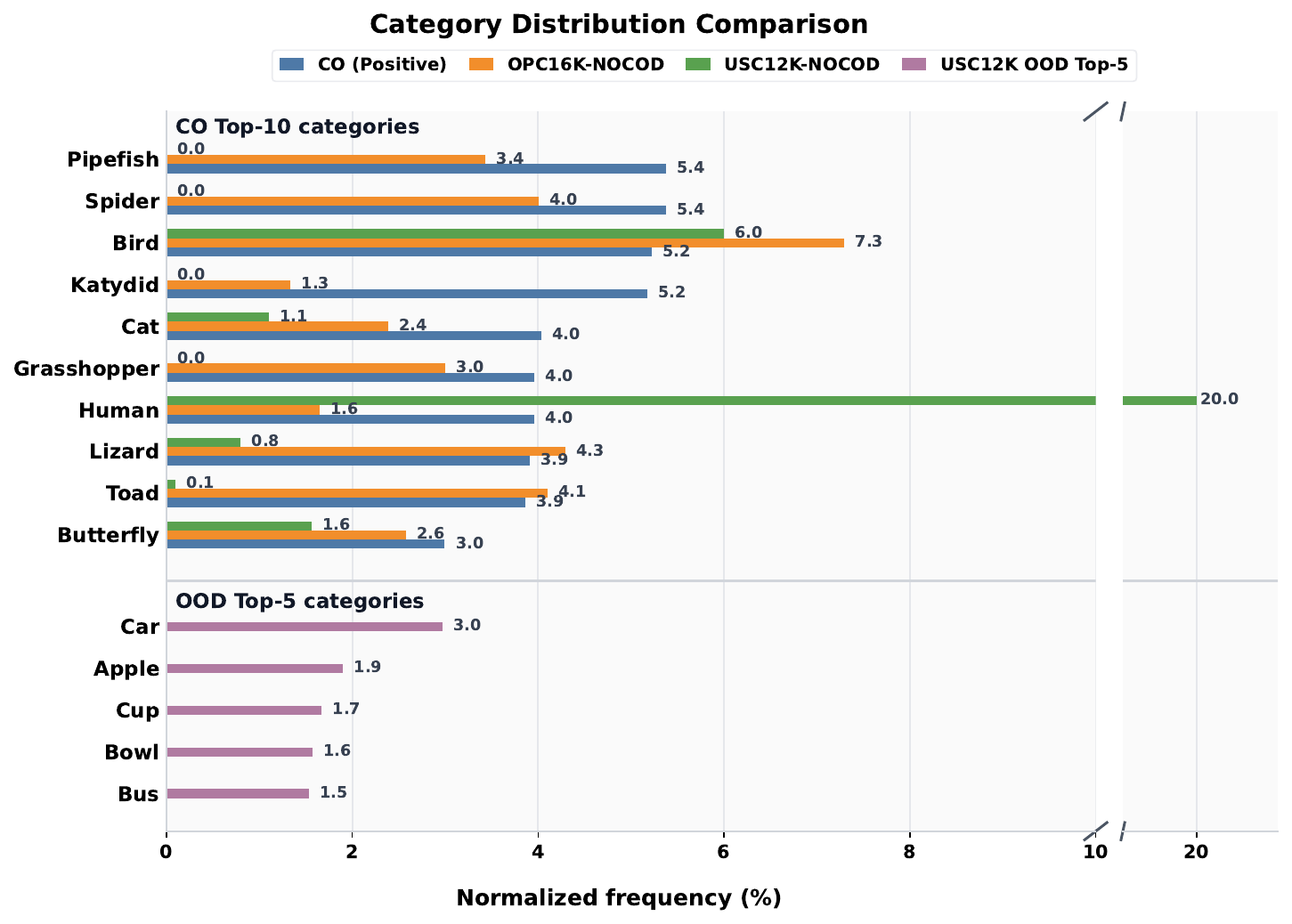}\\
\caption{Foreground-category distribution comparison among CO positives, OPC16K-NOCOD, and USC12K-NOCOD. The upper part shows the top-10 foreground categories ranked by their normalized frequencies in CO samples, while the lower part shows the top-5 out-of-domain categories observed in USC12K-NOCOD.}
\label{fig:2}
\vspace{-2mm}
\end{figure}

\subsection{Dataset Construction}
\label{sec:dataset}

OPC16K contains 16,245 images collected from 14 sources and is manually curated into three subsets: camouflaged-object images (CO), pure background images (BG), and non-camouflaged-object images (NOCOD), as illustrated in Fig.~\ref{fig:1}. Specifically, the CO subset consists of 9,000 images collected from existing COD datasets and web sources, with all samples manually screened to ensure genuine camouflage. The BG subset contains 3,050 object-free scene images selected from diverse background datasets, while the NOCOD subset contains 4,195 images featuring salient or general non-camouflaged objects. 
Together, BG and NOCOD serve as negative samples to evaluate false-positive suppression under realistic COD scenarios. Detailed source-wise statistics are provided in the supplementary material. To prevent data leakage, exact and near-duplicate images are removed through image hashing and perceptual similarity filtering prior to dataset splitting. For annotation, CO samples retain their original camouflaged-object masks when available. BG images are assigned all-background masks. For NOCOD samples, images filtered from existing COD datasets retain their original object masks after removing camouflaged instances, while the remaining samples are first annotated using SAM3~\cite{carion2025sam} and manually refined for quality assurance. OPC16K is split into 8,035 training images and 8,210 test images. The training set contains 3,790 CO, 2,050 BG, and 2,195 NOCOD images, while the test set contains 5,210 CO, 1,000 BG, and 2,000 NOCOD images. The CO split follows established protocols in prior COD works~\cite{fan2021concealed,pang2022zoom}, whereas the BG and NOCOD subsets are randomly partitioned from the negative samples.

\subsection{Distribution Alignment}

A key distinction of OPC16K is its deliberate consideration of foreground-category and scene-level distributions between positive and negative samples, which reduces potential category- and domain-level shortcuts in realistic COD evaluation.
For example, some salient-object negatives in USC12K~\cite{zhou2025rethinking} come from categories uncommon in standard COD benchmarks, such as indoor furniture and sports balls. To mitigate such category mismatch, we adopt a two-round collection process guided by the foreground-category distribution of camouflaged positive samples.
In the \textbf{first round}, we categorize the foreground classes of positive samples and rank them according to their occurrence frequencies. For example, categories such as \textit{spider} and \textit{insect} appear relatively frequently in camouflaged-object images, whereas categories such as \textit{chicken} and \textit{sheep} are less common. Using this distribution as a reference, we perform an initial collection pass for negative samples, aiming to obtain broad category coverage while avoiding a large deviation from the target distribution. 
In the \textbf{second round}, we compare the collected negative-sample distribution with the positive reference distribution and perform targeted supplementation for underrepresented categories. Some categories, such as snakes and insects, are difficult to obtain from general-purpose datasets, so we further collect samples from specialized sources, including SnakeCLEF~\cite{palaniappan2022deep} and IP102~\cite{Wu2019Insect}, to reduce the distributional discrepancy. Fig.~\ref{fig:2} visualizes the foreground-category frequency distributions of positive and negative samples after this supplementation. Beyond foreground categories, we further analyze scene-level distributions among CO, BG, and NOCOD subsets to examine potential scene-related biases, with detailed results provided in the supplementary material.

\subsection{Dataset Statistics and Comparison} 
We compare OPC16K with existing COD benchmarks, including CAMO~\cite{le2019anabranch}, COD10K~\cite{fan2020camouflaged}, NC4K~\cite{lv2021simultaneously}, CHAMELEON, and USC12K~\cite{zhou2025rethinking}, as summarized in Tab.\ref{tab:benchmark_comparison}. We introduce two properties for comparison: \textbf{Distribution Alignment (DA)}, indicating whether negative samples are deliberately collected according to the foreground-category distribution of positive samples; and \textbf{Out-of-Domain Negatives (OD)}, indicating whether the NOCOD subset contains object categories outside the COD domain. Most existing COD datasets, including CAMO, CHAMELEON, and NC4K, primarily focus on camouflaged positive samples and do not provide negative samples for realistic evaluation. COD10K includes background and non-camouflaged images, but these samples lack corresponding annotations and are therefore excluded from training and evaluation. USC12K introduces negative samples through salient objects; however, its NOCOD subset mainly consists of salient-object categories and does not cover general non-camouflaged objects. Moreover, a considerable proportion of its negative samples come from categories rarely observed in COD benchmarks, leading to category-level distribution mismatch between positive and negative samples. In contrast, OPC16K incorporates both salient and general non-camouflaged objects while maintaining foreground-category consistency with camouflaged positive samples. 
This design enables a more reliable evaluation of COD models by reducing potential biases introduced by foreground-category and scene-level distribution discrepancies.

\begin{figure*}[t]
\centering
% \vspace{-0.2em}
\includegraphics[width=1.0\linewidth]{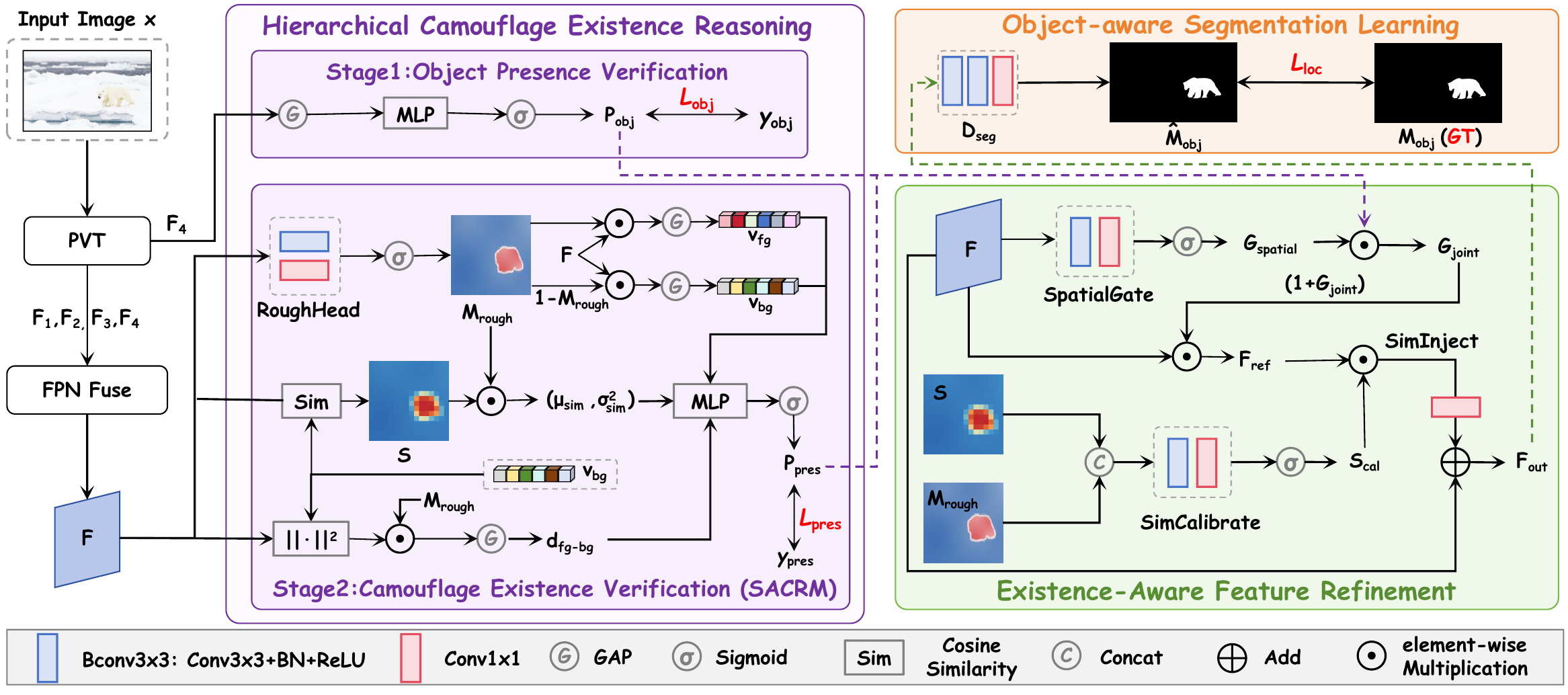}\\
\caption{Framework of OPCNet. OPCNet decouples COD into object localization and camouflage existence reasoning for realistic scenarios. The object-aware segmentation branch predicts category-agnostic object regions, while the hierarchical reasoning branch verifies object presence and camouflage existence via SACRM. The existence-aware refinement module then injects existence and similarity cues to generate the final COD mask.}

\label{fig:framwork}
\vspace{-2mm}
\end{figure*}

\section{Method}

\subsection{Overview}

Given an input image $\mathbf{x}\in\mathbb{R}^{H\times W\times3}$, OPCNet aims to identify whether a camouflaged object exists and generate accurate masks under realistic COD scenarios. Different from conventional COD methods that assume the existence of camouflaged objects and directly perform binary segmentation, we decouple COD into two complementary tasks: \textit{object localization} and \textit{camouflage existence reasoning}. The former determines where potential objects exist, while the latter identifies whether these objects exhibit camouflage characteristics. As shown in Fig.~\ref{fig:framwork}, OPCNet consists of three main components: 1) an object-aware segmentation branch that learns category-agnostic object localization from COD, NOCOD, and BG samples, 2) a hierarchical existence reasoning branch that progressively determines object presence and camouflage existence, and 3) an existence-aware feature refinement module that calibrates segmentation responses in realistic COD scenarios.

\subsection{Object-aware Segmentation Learning}

Unlike conventional COD methods that learn exclusively from camouflaged-object annotations, OPCNet leverages diverse supervision from COD, NOCOD, and BG samples to learn general object localization. Specifically, COD, NOCOD, and BG samples are supervised with camouflage masks, object masks, and all-zero masks, respectively. Given the fused feature representation $\mathbf{F}$, the localization decoder produces:
\begin{equation}
\hat{\mathbf{M}}_{obj}=D_{seg}(\mathbf{F}),
\end{equation}
where $\hat{\mathbf{M}}_{obj}$ represents the predicted object region. This branch focuses on learning \textit{where objects exist}, while camouflage recognition is handled by the existence reasoning branch.

\subsection{Hierarchical Camouflage Existence Reasoning}

To explicitly distinguish COD, NOCOD, and BG scenarios, OPCNet introduces a two-stage existence reasoning branch.

\noindent\textbf{Stage 1: Object Presence Verification.} Given the deepest backbone feature $\mathbf{F}_4$, we first predict whether an object exists in the input image. Global average pooling followed by an MLP classifier produces an object presence logit:
\begin{equation}
p_{obj}
=
\sigma
(
MLP_{obj}(GAP(\mathbf{F}_4))
),
\end{equation}
where $\sigma(\cdot)$ denotes the sigmoid activation function. This stage separates BG samples from object-containing images.

\noindent\textbf{Stage 2: Camouflage Existence Verification.} 
% For object-containing images, we further determine whether the object exhibits camouflage characteristics.
For object-containing images, we further determine whether the object exhibits camouflage characteristics. The camouflage existence branch is supervised only on COD and NOCOD samples, since BG inputs are already handled by the object presence verification in Stage 1. To distinguish camouflaged objects from non-camouflaged ones, this stage focuses on foreground-background relation modeling, motivated by the high similarity between camouflaged objects and their backgrounds.

\noindent\textbf{Similarity-Aware Camouflage Relation Module (SACRM).}
As the core component of Stage 2, SACRM models foreground-background relationships to infer camouflage existence. Given the fused feature $\mathbf{F}$, a lightweight rough localization head is applied to generate a soft object prior:
\begin{equation}
\mathbf{M}_{rough}
=
\sigma
(
\mathrm{RoughHead}(\mathbf{F})
),
\end{equation}
where $\mathrm{RoughHead}(\cdot)$ denotes a lightweight convolutional prediction head, and $\mathbf{M}_{rough}$ provides a soft localization prior to extract foreground and background prototypes.

Based on this prior, foreground and background prototypes are extracted through soft masked average pooling:
\begin{equation}
\mathbf{v}_{fg}
=
\frac{
\sum_p\mathbf{F}_p\mathbf{M}_{rough,p}
}
{
\sum_p\mathbf{M}_{rough,p}+\epsilon
},
\end{equation}

\begin{equation}
\mathbf{v}_{bg}
=
\frac{
\sum_p\mathbf{F}_p(1-\mathbf{M}_{rough,p})
}
{
\sum_p(1-\mathbf{M}_{rough,p})+\epsilon
}.
\end{equation}
where $\mathbf{F}_p$ denotes the feature vector at spatial location $p$, and $\epsilon$ is a small constant for numerical stability.

The similarity between each feature location and the background prototype is computed as:
\begin{equation}
\mathbf{S}_p
=
\hat{\mathbf{F}}_p\cdot\hat{\mathbf{v}}_{bg},
\end{equation}
where both $\mathbf{F}_p$ and $\mathbf{v}_{bg}$ are $\ell_2$ normalized. 
We summarize the similarity distribution within the predicted object region by computing its first- and second-order statistics, denoted as $\mu_{sim}$ and $\sigma^2_{sim}$. In addition, the foreground-background discrepancy is calculated as:
\begin{equation}
d_{fg\text{-}bg}
=
\frac{
\sum_p
||\mathbf{F}_p-\mathbf{v}_{bg}||_2^2
\mathbf{M}_{rough,p}
}
{
\sum_p\mathbf{M}_{rough,p}+\epsilon
}.
\end{equation}

These descriptors jointly characterize foreground representation, foreground-background similarity, and feature discrepancy. The camouflage existence probability is obtained from the resulting camouflage-aware relational descriptor:
\begin{equation}
p_{pres}
=
\sigma
\left(
\mathrm{MLP}_{pres}
(
[
\mathbf{v}_{fg};
\mathbf{v}_{bg};
\mathbf{v}_{fg}\text{-}\mathbf{v}_{bg};
\mu_{sim};
\sigma^2_{sim};
d_{fg\text{-}bg}
]
)
\right).
\end{equation}

\begin{table*}[!t]
\centering
\caption{Quantitative comparison on the OPC16K test set. We report three-way classification accuracy, negative-sample FPR, realistic COD segmentation metrics, and category-agnostic localization metrics. Best results are highlighted in \textbf{bold}.}
\label{tab:main_results}
\vspace{-1mm}
\resizebox{\linewidth}{!}{
\begin{tabular}{lc|cccc|cc|cccc|cccc}
\toprule
Method&Backbones&Acc$_{co}$$\uparrow$&Acc$_{bg}$$\uparrow$&Acc$_{no}$$\uparrow$& Acc$_{all}$$\uparrow$&FPR$_{bg}$ $\downarrow$&FPR$_{no}$$\downarrow$& oMAE $\downarrow$  &oS$_m$ $\uparrow$ & oE$_m$ $\uparrow$ &oF$^w_\beta$ $\uparrow$&coMAE $\downarrow$  &coS$_m$ $\uparrow$ & coE$_m$ $\uparrow$ &coF$^w_\beta$ $\uparrow$ \\
\midrule
SINet~\cite{fan2020camouflaged}  &ResNet50       & 0.814 & 0.853 & 0.798 & 0.814 & 0.142 & 0.196 & 0.224 & 0.659&0.674&0.537&0.051&0.785&0.829&0.583 \\
VSCode~\cite{luo2024vscode}&Swin-B&0.856&0.909&0.907&0.875&0.074&0.093&0.186&0.713&0.738&0.596&0.044&0.841&0.862&0.705\\
RUN~\cite{he2025run}      &PVTv2-B4     & 0.892 & 0.918 & 0.781 &0.868& 0.076 & 0.218 & 0.150 & 0.758 & 0.790&0.625&0.045&0.841&0.872&0.687 \\
CamoDiffusion~\cite{sun2025conditional} &PVTv2-B4& 0.909 & 0.903 & 0.815 & 0.884 & 0.084 & 0.182 & 0.135 & 0.780&0.786&0.654&0.042&0.858&0.860&0.715 \\
USCNet~\cite{zhou2025rethinking}&SAM-H        & 0.905& 0.809 & 0.679 &0.838& 0.179 & 0.321 & 0.149 & 0.758 & \textbf{0.885}&0.575&0.055&0.829&\textbf{0.976}&0.620 \\
\midrule
OPCNet (Ours)    &PVTv2-B4     & \textbf{0.930} & \textbf{0.949} & \textbf{0.817} & \textbf{0.905} & \textbf{0.050} & \textbf{0.181} & \textbf{0.105} & \textbf{0.818}&0.875&\textbf{0.701}&\textbf{0.038}&\textbf{0.874}&0.933&\textbf{0.745} \\
\bottomrule
\vspace{-6mm}
\end{tabular}}
\end{table*}

The final three-way prediction is obtained through hierarchical decision:

\begin{equation}
\hat{y}
=
\begin{cases}
\mathrm{BG},
&
p_{obj}\leq\tau_{obj},\\
\mathrm{COD},
&
p_{obj}>\tau_{obj},p_{pres}>\tau_{pres},\\
\mathrm{NOCOD},
&
p_{obj}>\tau_{obj},p_{pres}\leq\tau_{pres},
\end{cases}
\end{equation}
where $\tau_{obj}$ and $\tau_{pres}$ denote the decision thresholds for object presence and camouflage existence, respectively.

\subsection{Existence-Aware Feature Refinement }
\label{sec:gating}

The object localization branch learns general object cues from diverse scenarios but lacks explicit awareness of camouflage existence. To bridge the gap between object localization and camouflage existence reasoning, we introduce an existence-aware feature refinement module, which incorporates the predicted object presence and camouflage existence probabilities to adaptively calibrate segmentation features. A spatial gate is first generated from the fused feature representation:
\begin{equation}
\mathbf G_{spatial}
=
\sigma(\mathrm{SpatialGate}(\mathbf F)),
\end{equation}
where $\mathrm{SpatialGate}(\cdot)$ denotes a lightweight convolutional attention layer that produces spatially adaptive weights.

The spatial gate is further modulated by the hierarchical existence predictions:
\begin{equation}
\mathbf G_{joint}
=
p_{obj}\cdot p_{pres}\cdot \mathbf G_{spatial},
\end{equation}
where $p_{obj}$ and $p_{pres}$ denote the object presence probability and camouflage existence probability, respectively. For BG samples, the low $p_{obj}$ naturally suppresses $G_{joint}$ without requiring additional supervision on $p_{pres}$, as the hierarchical structure ensures that camouflage reasoning only effectively contributes when an object is detected. The refined feature is obtained through existence-aware feature calibration:
\begin{equation}
\mathbf F_{ref}
=
\mathbf F\odot
(1+\mathbf G_{joint})/2.
\end{equation}

The normalized gate scales feature responses between 0.5 and 1, allowing low-confidence regions to be attenuated while preserving reliable features. Furthermore, the similarity response generated by SACRM is injected into the refined feature representation. Specifically, the similarity map is first calibrated with the rough localization prior:
\begin{equation}
\mathbf S_{cal}
=
\sigma(
\mathrm{SimCalibrate}
([\mathbf S;\mathbf M_{rough}])
),
\end{equation}
where $\mathrm{SimCalibrate}(\cdot)$ denotes a lightweight convolutional module that refines the similarity response using the rough localization prior. Then, the calibrated similarity response is integrated into the segmentation feature:
\begin{equation}
\mathbf F_{out}
=
\mathbf F_{ref}
+
\mathrm{SimInject}
(
\mathbf F_{ref}\odot\mathbf S_{cal}
),
\end{equation}
where $\mathrm{SimInject}(\cdot)$ projects the calibrated similarity response into the segmentation feature space. The final segmentation mask is generated from $\mathbf F_{out}$.

\subsection{Loss Function}

OPCNet is optimized with a multi-task objective:

\begin{equation}
\mathcal L
=
\mathcal L_{loc}
+
\mathcal L_{obj}
+
\mathcal L_{pres}.
\end{equation}

The localization loss is applied to all samples:
\begin{equation}
\mathcal L_{loc}
=
\frac{1}{N}
\sum_i
\mathcal L_{BCE}
(
\hat{\mathbf M}_i,
\mathbf M_i
),
\end{equation}
where $\mathbf M_i$ denotes the object localization target. 
% and $w_i$ balances the contribution of different sample types.
The object presence loss is defined as:
\begin{equation}
\mathcal L_{obj}
=
\mathrm{BCE}
(
p_{obj},
y_{obj}
),
\end{equation}
where $y_{obj}=1$ for COD and NOCOD samples and $y_{obj}=0$ for BG samples. The camouflage existence loss is optimized on object-containing samples:
\begin{equation}
\mathcal L_{pres}
=
\frac{1}{|\mathcal B|}
\sum_i
\mathrm{BCE}
(
p_{pres,i},
y_{pres,i}
),
\end{equation}
where $y_{pres}$ denotes the camouflage-existence label.

\section{Experiments}

\subsection{Experimental Setup}

\noindent\textbf{Dataset.} We evaluate all methods on OPC16K. The dataset is split into 8,035 training images (CO: 3,790; BG: 2,050; NOCOD: 2,195) and 8,210 test images (CO: 5,210; BG: 1,000; NOCOD: 2,000). The camouflaged-object split follows the established settings of prior COD works~\cite{fan2021concealed}, while BG and NOCOD samples are randomly partitioned from the collected negative samples.

\noindent\textbf{Implementation Details.}
OPCNet is implemented in PyTorch and trained on a single NVIDIA RTX A6000 GPU. The backbone adopts PVTv2-B4~\cite{wang2022pvt}. All input images are resized to $512 \times 512$. We use AdamW as the optimizer, with a backbone learning rate of $1\times10^{-5}$, a head learning rate of $1\times10^{-4}$, weight decay of $1\times10^{-4}$, and a cosine annealing scheduler over 30 epochs. During inference, the object-presence and camouflage-existence thresholds are both set to 0.5 following the default probability decision rule.

\noindent\textbf{Compared Methods.}
We compare OPCNet with five COD methods: SINet~\cite{fan2020camouflaged}, VSCode~\cite{luo2024vscode}, RUN~\cite{he2025run}, CamoDiffusion~\cite{sun2025conditional}, and USCNet~\cite{zhou2025rethinking}. For fair comparison, all methods are retrained on OPC16K using their official implementations and recommended training settings. Since existing COD methods are originally designed for binary camouflage segmentation, we adapt them to the proposed realistic COD setting by extending their prediction space to three classes: CO, NOCOD, and BG. All baselines are trained with the same pixel-level supervision protocol as OPCNet. Detailed adaptation strategies are provided in the appendix.

\subsection{Evaluation Protocol}

Conventional COD metrics, including MAE, S$_m$~\cite{fan2017structure}, E$_m$~\cite{fan2018enhanced}, and F$^w_\beta$, are designed for positive-only evaluation where camouflaged objects are assumed to exist. In realistic COD evaluation, however, models should not only segment COD samples but also reject BG and NOCOD inputs. We therefore evaluate models from four aspects. The prefix “o” denotes overall evaluation across all subsets, while “co” denotes evaluation on the CO subset.

\noindent\textbf{Three-way Classification Accuracy.}
We report overall and per-category classification accuracies, denoted as $\mathrm{Acc}_{all}$, $\mathrm{Acc}_{co}$, $\mathrm{Acc}_{bg}$, and $\mathrm{Acc}_{no}$, for the three scenarios: CO, BG, and NOCOD. Since existing COD methods produce pixel-level masks, we convert their predictions into image-level labels by assigning BG when all pixels are predicted as background and selecting the category with the largest foreground area otherwise. These metrics evaluate the capability of recognizing different realistic COD scenarios.

\noindent\textbf{False Positive Rate on Negative Samples.}
Since CO is the target class, false alarms on negative samples are particularly harmful. We report the CO false positive rates on BG and NOCOD subsets, denoted as $\mathrm{FPR}_{bg}$ and $\mathrm{FPR}_{no}$, respectively.

\noindent\textbf{Overall COD Segmentation Quality.}
For COD samples, we introduce overall segmentation metrics, denoted as oMAE, oS$_m$, oE$_m$, and oF$^w_\beta$. Correctly classified COD samples are evaluated with standard segmentation metrics, while misclassified samples are assigned the worst scores, e.g., MAE = 1 and S$_m$ = 0. These metrics jointly evaluate COD recognition and segmentation quality under realistic COD scenarios.

\begin{figure}[t]
\centering
\includegraphics[width=\linewidth]{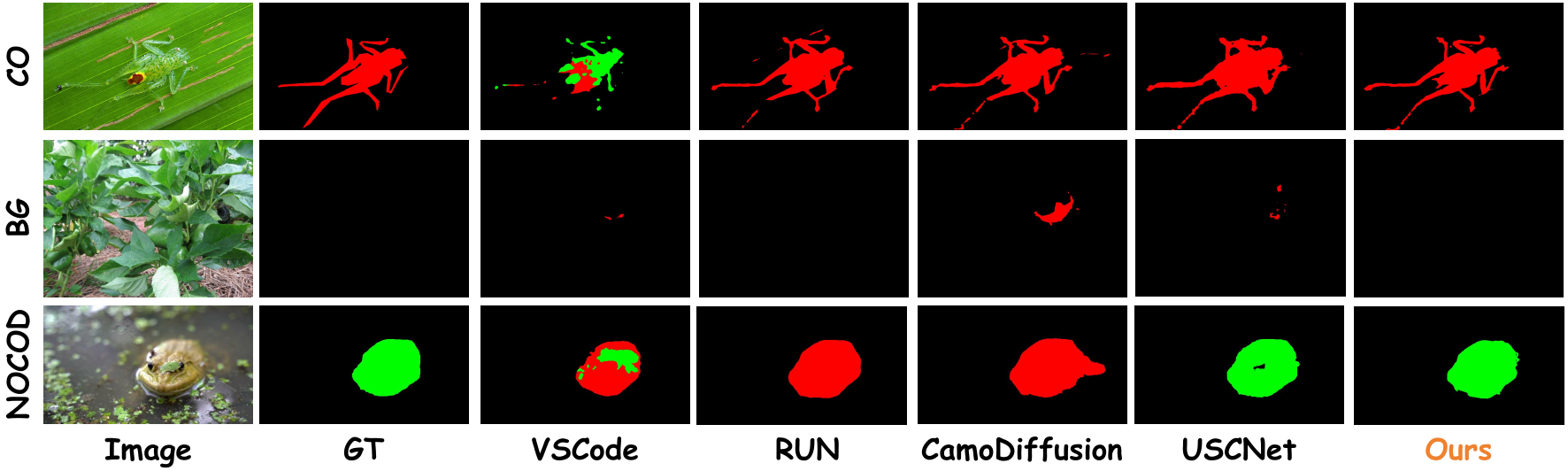}\\

\caption{Qualitative comparison on representative CO, BG, and NOCOD samples from OPC16K.}\label{fig:visual}
\vspace{-2mm}
\end{figure}

\noindent\textbf{CO Segmentation Quality.}
To evaluate the segmentation capability on camouflaged objects independently of scene recognition, we compute standard segmentation metrics on CO samples without considering predicted scene labels. The corresponding metrics are denoted as coMAE, coS$_m$, coE$_m$, and coF$^w_\beta$. Detailed definitions are provided in the appendix.

\subsection{Main Results}

\noindent\textbf{Quantitative Comparison.}
Tab.~\ref{tab:main_results} shows the comprehensive comparison on the OPC16K test set. OPCNet achieves the best overall performance under the proposed open-world COD protocol. It obtains the highest three-way classification accuracy and the lowest $\mathrm{FPR}_{bg}$ and $\mathrm{FPR}_{no}$, demonstrating the effectiveness of hierarchical existence reasoning in distinguishing CO, BG, and NOCOD scenarios. Compared with the original models reported in Tab.~\ref{tab:1}, retraining existing methods on OPC16K improves their negative-sample rejection, but they still underperform OPCNet, indicating the necessity of explicit existence modeling. For segmentation, OPCNet achieves superior open-world COD metrics, including oMAE, oS$_m$, oE$_m$, and oF$_\beta^w$, while maintaining competitive category-agnostic localization performance. These results indicate that explicit existence modeling improves open-world reliability without compromising object localization quality.

\noindent\textbf{Qualitative Comparison.}
Fig.~\ref{fig:visual} shows visual comparisons across CO, BG, and NOCOD scenarios. Existing methods often activate on background patterns or non-camouflaged objects, producing false COD masks on negative samples. OPCNet effectively rejects these negative cases through hierarchical existence reasoning and existence-aware feature refinement. Meanwhile, on true CO samples, OPCNet generates more complete and boundary-preserving masks, indicating that explicit existence modeling improves open-world reliability without compromising segmentation quality.

% \begin{table}[t]
% \centering
% \caption{Ablation study of key components in OPCNet.}
% \label{tab:ablation}
% \vspace{-1mm}

% \resizebox{\linewidth}{!}{
% \begin{tabular}{cccc|ccc|cccc}
% \toprule
% \textbf{Base} & \textbf{HER} & \textbf{SACRM} & \textbf{EAFR}
% &
% \textbf{Acc$_{all}$}$\uparrow$
% &
% \textbf{FPR$_{bg}$}$\downarrow$
% &
% \textbf{FPR$_{no}$}$\downarrow$
% &
% \textbf{oMAE}$\downarrow$
% &
% \textbf{oS$_m$}$\uparrow$
% &
% \textbf{coMAE}$\downarrow$
% &
% \textbf{coS$_m$}$\uparrow$
% \\
% \midrule
% \checkmark &  &  & 
% & 0.771 & 0.083 & 0.208 & 0.304 & 0.616 & 0.046 & 0.831 \\
% \checkmark & \checkmark &  & 
% & 0.872 & 0.057 & 0.189 & 0.157 & 0.761 & 0.045 & 0.837 \\
% \checkmark & \checkmark & \checkmark & 
% & 0.901 & 0.053 & 0.183 & 0.136 & 0.784 & 0.041 & 0.852 \\
% \checkmark & \checkmark & \checkmark & \checkmark
% & \textbf{0.905} & \textbf{0.050} & \textbf{0.181} & \textbf{0.105} & \textbf{0.818} & \textbf{0.038} & \textbf{0.874} \\
% \bottomrule
% \vspace{-4mm}
% \end{tabular}
% }

% \end{table}

\begin{table}[!t]
\centering
\caption{Ablation study of key components in OPCNet.}
\label{tab:ablation}
\vspace{-1mm}
\setlength{\tabcolsep}{2pt}
\resizebox{\linewidth}{!}{
\begin{tabular}{cccc|ccc|cccc}
\toprule
\textbf{Base} & \textbf{HER} & \textbf{SACRM} & \textbf{EAFR}
&
\textbf{Acc$_{all}$}$\uparrow$
&
\textbf{FPR$_{bg}$}$\downarrow$
&
\textbf{FPR$_{no}$}$\downarrow$
&
\textbf{oMAE}$\downarrow$
&
\textbf{oS$_m$}$\uparrow$
&
\textbf{coMAE}$\downarrow$
&
\textbf{coS$_m$}$\uparrow$
\\
\midrule
\checkmark &  &  & 
& 0.771 & 0.083 & 0.208 & 0.304 & 0.616 & 0.046 & 0.831 \\
\checkmark & \checkmark &  & 
& 0.872 & 0.057 & 0.189 & 0.157 & 0.761 & 0.045 & 0.837 \\
\checkmark & \checkmark & \checkmark & 
& 0.901 & 0.053 & 0.183 & 0.136 & 0.784 & 0.041 & 0.852 \\
\checkmark & \checkmark & \checkmark & \checkmark
& \textbf{0.905} & \textbf{0.050} & \textbf{0.181} & \textbf{0.105} & \textbf{0.818} & \textbf{0.038} & \textbf{0.874} \\
\bottomrule
\vspace{-4mm}
\end{tabular}
}
\end{table}

\subsection{Ablation Study}

To verify the effectiveness of each component in OPCNet, we conduct ablation studies on the OPC16K test set. The results are reported in Tab.~\ref{tab:ablation}. Here, Base denotes the object-aware segmentation branch without explicit existence reasoning.
\noindent\textbf{Effect of HER.}
% HER provides substantial improvements in $\mathrm{Acc}_{all}$ and negative-sample FPRs, validating the effectiveness of explicit object presence and camouflage existence reasoning.
Introducing HER substantially improves $\mathrm{Acc}_{all}$ and reduces both $\mathrm{FPR}_{bg}$ and $\mathrm{FPR}_{no}$, demonstrating the effectiveness of decomposing COD into object presence verification and camouflage existence reasoning.

\noindent\textbf{Effect of SACRM.}
Adding the Similarity-Aware Camouflage Relation Module (SACRM) further improves overall COD performance by modeling foreground-background similarity and discrepancy, providing explicit relational cues for camouflage existence reasoning.

\noindent\textbf{Effect of EAFR.}
% EAFR brings additional gains on overall and CO-only segmentation quality by calibrating segmentation features with existence cues.
EAFR brings additional gains on overall and CO-only segmentation quality by calibrating segmentation features with existence cues.

\section{Conclusion}

In this paper, we study camouflaged object detection under realistic scenarios, where inputs may contain camouflaged objects, non-camouflaged objects, or pure backgrounds. Different from conventional COD settings that assume the existence of camouflaged objects, we highlight the importance of reliable object existence reasoning and false-positive suppression. To this end, we construct OPC16K, a large-scale benchmark for realistic COD evaluation, and propose OPCNet, which decouples COD into object localization and camouflage existence reasoning through hierarchical reasoning and existence-aware feature refinement. Extensive experiments demonstrate that OPCNet achieves superior performance by reducing false positives on negative samples while maintaining accurate segmentation of camouflaged objects. We hope OPC16K and OPCNet can facilitate future research on reliable COD in practical scenarios.

\bibliography{aaai2027}


\end{document}